\documentclass{article}

\usepackage{PRIMEarxiv}

\usepackage[accsupp]{axessibility}  

\usepackage{multirow}

\usepackage[utf8]{inputenc} 
\usepackage[T1]{fontenc}    
\usepackage{hyperref}       
\usepackage{url}            
\usepackage{booktabs}       
\usepackage{amsfonts}       
\usepackage{nicefrac}       
\usepackage{microtype}      
\usepackage{lipsum}
\usepackage{fancyhdr}       
\usepackage{graphicx}       

\usepackage[table,xcdraw]{xcolor}

\usepackage[capitalize]{cleveref}
\crefname{section}{Sec.}{Secs.}
\Crefname{section}{Section}{Sections}
\Crefname{table}{Table}{Tables}
\crefname{table}{Tab.}{Tabs.}

\title{3D Adaptive Structural Convolution Network for Domain-Invariant Point Cloud Recognition 
}

\author{
 Younggun Kim\thanks{He is currently with University of Central Florida}, Beomsik Cho, Seonghoon Ryoo, and Soomok Lee\thanks{Corresponding author.} \\
  Department of Mobility Engineering \\
  Ajou University, Suwon, Korea, 16499\\
  \texttt{\{kyg9191, whqjatlr123, shryu0512, soomoklee\}@ajou.ac.kr 
} \\
}
\begin{document}
\maketitle

\begin{abstract}
 Adapting deep learning networks for point cloud data recognition in self-driving vehicles faces challenges due to the variability in datasets and sensor technologies, emphasizing the need for adaptive techniques to maintain accuracy across different conditions. In this paper, we introduce the 3D Adaptive Structural Convolution Network (3D-ASCN), a cutting-edge framework for 3D point cloud recognition. It combines 3D convolution kernels, a structural tree structure, and adaptive neighborhood sampling for effective geometric feature extraction. This method obtains domain-invariant features and demonstrates robust, adaptable performance on a variety of point cloud datasets, ensuring compatibility across diverse sensor configurations without the need for parameter adjustments. This highlights its potential to significantly enhance the reliability and efficiency of self-driving vehicle technology.
 \end{abstract}


\section{INTRODUCTION}

Self-driving vehicles \cite{review1, review2} have ushered in a new era of transportation, promising to improve safety, efficiency, and convenience with their advanced capabilities. Central to their operation is the accurate perception of the surrounding environment, a task crucially supported by Light Detection and Ranging (LiDAR) sensors. Distinct from cameras and radar, LiDAR excels by providing high-precision 3D spatial data. Operating across various channels, such as 16, 32, or 64, and rotating 360 degrees, LiDAR is able to create a comprehensive point cloud of obstacles over substantial distances, a capability that is crucial for navigation in urban areas. This feature is particularly beneficial for the effective detection of 3D objects and navigating complex environments filled with numerous objects and distinct features.

\begin{figure*}[!t]
    \centerline{\includegraphics[width=\textwidth]{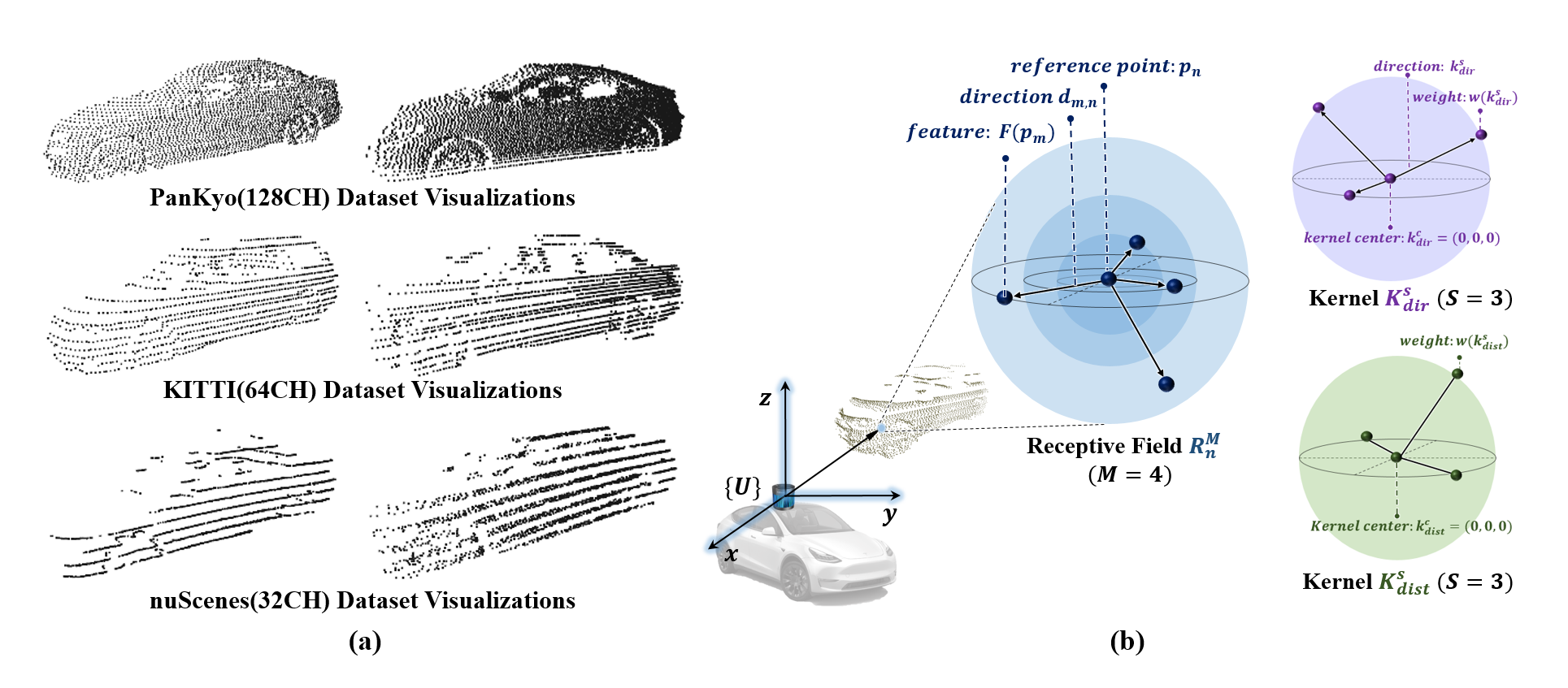}}
    \caption{Our proposed convolution concept and its target for different point cloud domain. (a) illustrates point cloud visualizations in different source domains. (b) shows receptive field, direction-based kernel, and distance-based kernel.}
    \label{Fig1} 
\end{figure*}

Adapting deep learning networks for point cloud data recognition introduces distinct challenges, especially with domain variability illustrated in \cref{Fig1} (a), such as differences in dataset characteristics or sensor manufacturers. Research in this area primarily explores how diversity in datasets affects deep learning models. It is observed that models trained on point cloud data under one set of conditions or with a specific sensor often face performance issues when applied to data from different conditions or sensors. For instance, a shift from training on high-density point cloud data to testing on lower-density data, due to cost considerations, can lead to a noticeable drop in model performance. This scenario highlights the sensitivity of deep learning models to the specific attributes of point cloud data, emphasizing the necessity for adaptive techniques capable of handling variations in data domains and acquisition methods with different platforms or sensor suites.

PointNet \cite{pointnet}, the first network to encode point cloud data, revolutionized its treatment by organizing it into one-dimensional arrays for processing through a multi-layer perceptron (MLP). This approach preserves the data's unstructured nature via max-pooling across MLP layers without converting it to structured formats. However, despite its pioneering method and improvements by PointNet++ \cite{pointnet+}, it overlooks the inherent 3D structure of point cloud data. Instead, it uses a rotation matrix to approximate spatial relationships, which can lead to overfitting and struggles with different sensor setups. This limitation hampers its ability to handle the dynamic nature of point cloud data across various environments. Additionally, current encoding methods, including PointNet’s, often fail to capture the full 3D context, simplifying environmental variations such as different vegetation types or road configurations.

Meanwhile, graph-based methods such as Dynamic Graph CNN (DGCNN) \cite{dgcnn} can explain 3D geometric information more deeply, utilizing graph-based convolution networks alongside a k-nearest neighbor strategy to form dynamic graphs that enhance local feature understanding. This approach facilitates a nuanced comprehension of point-to-point relationships, contributing to more sophisticated feature extraction. However, DGCNN's complexity introduces computational challenges and a dependency on the precision of input graphs, where suboptimal graph constructions can undermine the outcomes. Addressing these complexities requires innovative solutions to ensure rigid and comprehensive feature extraction.

In this paper, we present a 3D Adaptive Structural Convolution Network (3D-ASCN) as the optimal method for encoding in LiDAR-based 3D recognition tasks within the self-driving dataset. By electing optimal neighborhoods of each point, applying 3D convolution kernels in combination with a 3D geometric tree structure, and utilizing cosine similarity and Euclidean distance, our approach effectively extracts geometric features that are more appropriate and maintains consistency in features extracted from LiDAR data, irrespective of regional variations or changes in LiDAR sensors. Through our proposed method of structural encoding, we have achieved significant improvements in performance across different domains, including changes in datasets or LiDAR sensor configurations. The main contributions of this study can be summarized in three folds as follows:

\begin{itemize}

\item We propose novel LiDAR-invariant 3D convolution kernels to train 3D structural perspective by combining Cosine similarity and Euclidean distance term. Our network shows stability and consistent performance in varying point cloud datasets.

\item An adaptive neighborhood sampling method is proposed, based on principal components of the 3D covariance ellipsoid, which allows us to exploit highly notable geometric features by selecting the optimal neighborhood numbers.

\item The proposed 3D-ASCN method demonstrates domain-invariant feature extraction across diverse types of LiDAR data and platforms. The  robust performance is demonstrated with a variety of real-world point cloud datasets, showcasing the structural features' domain invariance. When tested on high-resolution point clouds from vastly different areas and resolutions, the model maintains its highest robustness and adaptability to various sensor configurations without necessitating parameter adjustments.
\end{itemize}

The subsequent sections of this paper are organized as follows. Section \ref{section2} presents a comprehensive literature review of the point cloud encoding networks with respect to LIDAR object classification. Section \ref{section3} illustrates the proposed 3D-ASCN configuration and classification network architecture. In Section \ref{section4}, the paper demonstrates the evaluation of the proposed adaptation concerning varying LiDAR channels. Lastly, Section \ref{section5} concludes the paper.

\section{RELATED WORK}
\label{section2}

\subsection{ 3D point cloud-based Networks  }

\subsubsection{Pointwise MLP networks} PointNet \cite{pointnet} processes unordered 3D point cloud data using shared fully connected layers and channel-wise max-pooling to extract global features. However, it primarily learns key point representations and does not encode local geometric information, making it sensitive to variations in translation and scaling. To address this, researchers have proposed sorting 3D points into ordered lists where neighboring points have smaller Euclidean distances. For example, one study \cite{p8} sorts points along different dimensions and uses Recurrent Neural Networks (RNN) to extract features, while others \cite{p5, p10} use a kd-tree to convert 3D points into a 1D list followed by 1D CNNs. However, this approach can struggle to preserve local geometric information. PointMLP \cite{pointMLP} asserts that detailed local geometric information might not be essential for point cloud analysis. It utilizes a pure residual network without a complex local geometry extractor. Instead, it includes a lightweight geometric affine module, which significantly improves inference speed. PointMLP shows outstanding performance on both the ModelNet40 dataset \cite{ModelNet40}, which is ideal for CAD-based analysis, and the ScanObjectNN dataset \cite{ScanObjectNN}, which represents real-world indoor environments. PointNet++ \cite{pointnet+} based on a hierarchical feature learning paradigm to recursively capture local geometric structures demonstrates promising results and has become the cornerstone of modern point cloud methods due to its local point representation. Based on PointNet++, PointNeXt \cite{pointNext} improves the training and training strategies to enhance the performance of PointNet++ and introduces a separable MLP along with an inverted residual bottleneck design within the PointNet++ framework. PointNeXt surpasses state-of-the-art methods in 3D classification.

\subsubsection{Graph convolution networks} 

The Dynamic Graph CNN (DGCNN) \cite{dgcnn} represents a significant advancement in 3D point cloud processing with its graph-based convolutional networks. It uses a k-nearest neighbor algorithm to create a dynamic graph that captures local point-to-point relationships for improved feature extraction. However, DGCNN is computationally intensive due to the complexity of dynamic graph adjustments. Its performance also hinges on the quality of the input graph, where suboptimal or noisy graphs can affect results, and refining the graph construction process is complex. To improve graphical point cloud relationships in convolution, Lin {\it et al.} propose a 3D Graph Convolutional Network \cite{gcn} to learn geometric properties of point clouds. Their approach stands out for considering geometric values and adjusting network parameters through directional analysis. The model uses deformable graph kernels to handle translation, scale, and z-axis rotation. However, its effectiveness is mainly evaluated on ideal CAD-based datasets, which do not fully represent realistic occlusion scenarios in point cloud data. Additionally, it does not adequately account for the distances between points, potentially missing detailed interval nuances during parameter training, which is crucial for self-driving datasets with consistent but small-scale patterns despite variations in LiDAR technology.
Furthermore, although both DGCNN and GCN adopt the number of the neighborhoods as a constant value when selecting neighborhoods, each point in the point cloud will probably require a different number of neighborhoods. Accordingly, not only may this approach be able to extract distorted local region features, but if the neighborhood is set to be large, the degree of distortion will increase as the network becomes more advanced. Therefore, the optimal number of neighborhoods must be selected for every point by variably adjusting the number of neighborhoods.

\subsection{ Adaptive neighborhood sampling }
To describe the 3D structure surrounding a point using geometric features, it is essential to define a local neighborhood that encompasses all considered 3D points. Various strategies exist for defining these local neighborhoods around a specific 3D point. The most commonly used neighborhood definitions include Spherical Neighborhood \cite{sphere}, Cylindrical Neighborhood \cite{cylinder}, and K-Nearest Neighborhood (K-NN) \cite{knearest}. Spherical neighborhood approach involves defining the neighborhood as all 3D points within a sphere of a predetermined radius centered on the point. Cylindrical neighborhood  method consists of all 3D points whose 2D projections fall within a circle of a fixed radius around the point's projection. The k-nearest neighbors definition uses a fixed number of the closest neighbors to the point in 3D space.

These methods, which rely on a constant scale parameter, offer a simple way to select neighborhoods. The parameter is either a fixed radius or a set number of neighbors. However, the choice of scale parameter often depends on heuristic or empirical knowledge across point cloud datasets. To circumvent making strong assumptions about the local 3D neighborhoods of each point, more recent studies have aimed at determining an optimal neighborhood size for each individual point, thereby enhancing the uniqueness of the derived features. Most approaches focus on optimizing the number of closest neighbors (k) for each point, which could be done through the methods that consider factors like curvature, point density, and noise in normal estimation. \cite{Scaleselection} These factors are especially pertinent for densely sampled, nearly continuous surfaces.

Other techniques consider variations in the surface \cite{curvature} or are based on the dimensionality \cite{dimensionality} or eigenentropy \cite{entropy} to select the optimal neighborhood size. While methods based on surface variation and dimensionality require heuristic judgment, the eigenentropy method can automatically determine the optimal neighborhood size. We further refined this eigenentropy approach to make it more suitable for application in our 3D-ASCN.

\section{3D ADAPTIVE STRUCTURAL CONVOLUTION NETWORK (3D-ASCN)}
\label{section3}

Our 3D adaptive structural convolution network (3D-ASCN) is specifically designed to process data captured by LIDAR point cloud sensors. This network ingests in 3D point cloud data and analyzes it to identify important geometric structures. Specifically, 3D-ASCN extracts 3D structural context features, including geometric information, ensuring performance does not depend on the point density of the dataset and enables domain-invariant point cloud recognition.

\subsection{Receptive Field for 3D-ASCN}

A 3D point cloud object is referred to as $P$, consisting of $N$ points represented by $P = ({p_{n}|n = 1, 2, \ldots, N})$. $p_{n}$ represents a 3D coordinate within the three-dimensional space, thus belonging to the set $\mathbb{R}^{3}$. For each point in the point cloud, a specific derived feature, $F(p)$, is associated, which is a vector of dimension $D$. This vector, $F(p) \in \mathbb{R}^{D}$, encapsulates certain characteristics or attributes of the point. In order to grasp the local structural and geometric features associated with each point $p_{n}$, we define the receptive field, denoted as ${R}^{M}_{n}$, which consists of $M$ neighborhood points. This receptive field for the point $p_n$, with a size of $M$, is established as shown in \cref{Fig1} (b) and defined as:

\begin{equation}
R^{M}_ {n} = \{p_{n}, p_{m} | \forall p_{m} \in \mathcal{N}(p_{n},M)\}
\label{eq1}
\end{equation}

where $\mathcal{N}(p_{n},M)$ formally represents the $M$ nearest neighbor points of $ p_{n}$, and the directional vectors $d_{m, n} = p_{m} - p_{n}$ will be used for later farthest neighborhood selection among points $p_{m}$ in a receptive field and for structural convolution purposes. The features encompassed within the input points $p_{n}$, with a size of $M$, can be expressed as $\{F (p_{n}), F (p_{m}) | \forall p_{m} \in \mathcal{N}(p_{n},M)\}$. These features are computed  during the structural convolution operation.

\subsection{3D Structural Kernels}

To perform convolution in 3D point cloud structures, we compose direction-based kernels and distance-based kernels. 
\subsubsection{Direction-based kernel} Similar to other 3D Graph Convolution  Kernel concepts \cite{gcn}, Kernel $K^S$ is computed for graph computation. We design it as a direction-based kernel, denoted as $K^{S}_{dir}$, where $S$ indicates the number of branches. More precisely, it consists of $S$ branches to train the directions and weights of the directional vectors. In order to define the direction, $K^{S}_{dir}$ requires the center kernel $k^{c}_{dir} = (0, 0, 0)$, from which each support includes $k^{1}_{dir}$, $k^{2}_{dir}$, \ldots, $k^{S}_{dir}$. $K^{S}_{dir}$ is composed of the center $k^{c}_{dir}$ and supports for the directional vectors $k_{j} \in \mathbb{R}^{3}$ and its combination $K^{S}_{dir} = \{k^{c}_{dir}, k^{1}_{dir}, k^{2}_{dir}, \ldots, k^{S}_{dir}\}$. Then, with direction-based weight vectors defined as $w(k_{dir}) \in \mathbb{R}^{D}$ for each kernel point $k_{dir}$, we achieve a part of the convolution operations through the weighted sum of features corresponding to the directional weights of $F(p)$. Since it is designed to train the structures in 3D, the directional vectors, represented by $k^{s}_{dir} - k^{c}_{dir} = k^{s}_{dir}$, are among the trained kernel parameters.

\subsubsection{Distance-based kernel} We propose a distance-based kernel, denoted as $K^{S}_{dist}$, to reflect the distance between each point as a learning parameter. Simillar to the direction kernel, $S$ denotes the number of branches, and each branch contains weights for the distance relationship between points, defined as $w(k_{dist}) \in \mathbb{R}^{D}$. Then, we perform a part of convolution operations via the weighted sum of features corresponding to the distance-based weights of $F(p)$. Therefore, with both the direction-based kernel and distance-based kernel included, the total trainable kernels in this network are defined as $\{w(k^{c}_{dir}), (k^{s}_{dir}, w(k^{s}_{dir})), w(k^{c}_{dist}), w(k^{s}_{dist}) \mid s = 1, 2, \ldots, S\}$.\\

\begin{figure*}[!t]
    \centerline{\includegraphics[width=\textwidth]{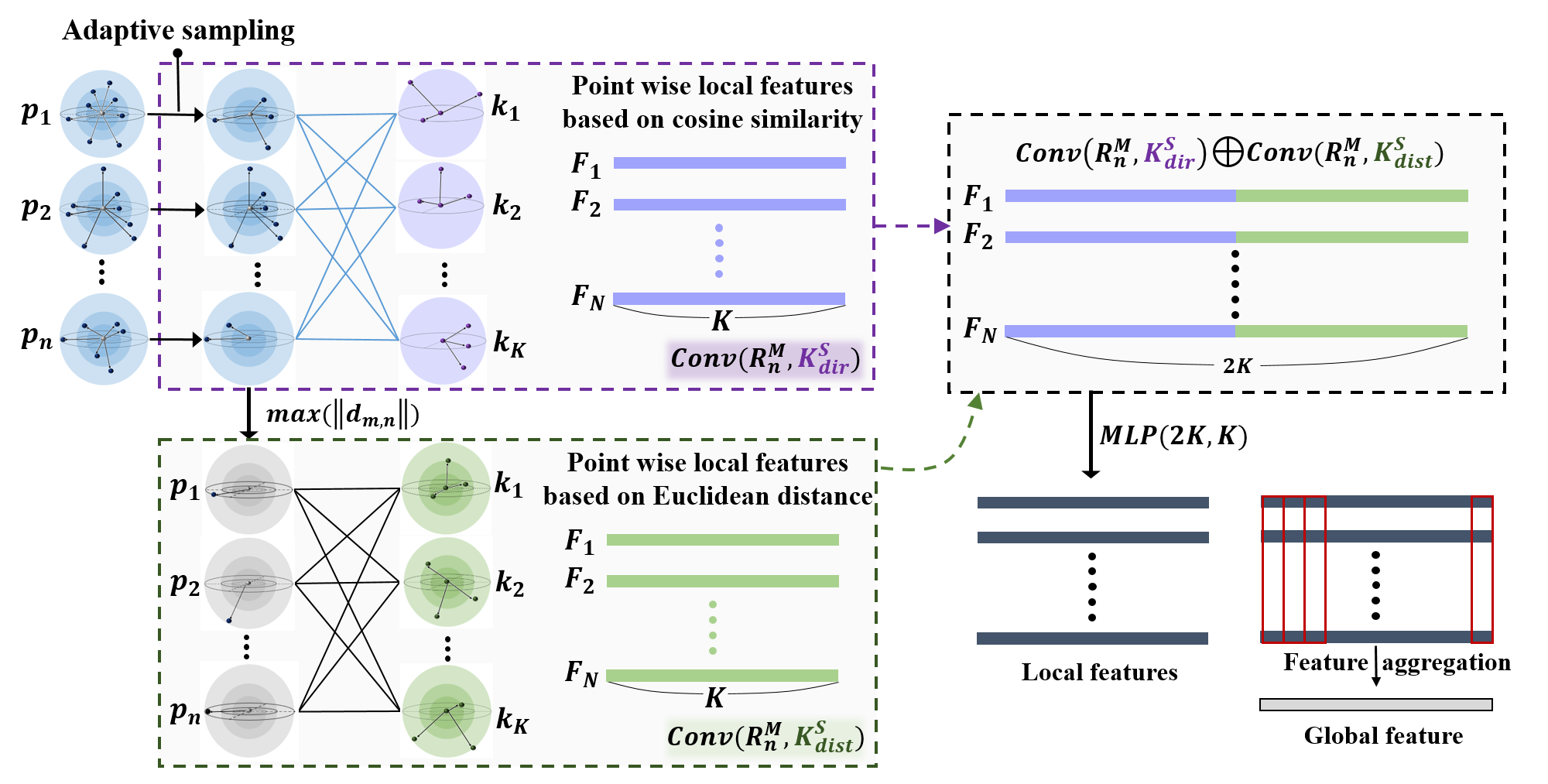}}
    \caption{Adaptive sampling and structural convolution operation: All points select their optimal neighborhoods based on k-nearest neighborhood selection and by minimizing Shannon entropy. Then, structural convolution operations are performed to extract neighborhood features of all points based on cosine similarity and Euclidean distance. Features based on cosine similarity and Euclidean distance are concatenated and passed through a multi-layer perceptron. These features are refined as they are used as inputs for this network. Finally, we can extract a global feature that includes structural information of a point cloud by aggregating all local features.}
    \label{Fig3} 
\end{figure*}

\subsection{Adaptive neighborhood sampling}
Since our network obtains local region geometric information through the distance and directional relationships between each point $p_{n}$ in the point cloud and the neighborhoods, all $p_{n}$ require different neighborhood sizes based on geometric properties. Additionally, the point density in the captured 3D point cloud data varies with the type of LiDAR sensor, and this density significantly influences the selection of the number of neighborhoods. Our 3D-ASCN is not only capable of extracting excellent geometric information from a point cloud by selecting the optimal number of neighborhood points for each point $p_{n}$, but it also remains invariant in LiDAR channel changes. \\

K-NN \cite{knearest} algorithm is applied to each point as the neighborhood search method for $p_{n}$, allowing more flexibility concerning the geometric size of the neighborhood. Instead of arbitrarily choosing a fixed number for the receptive field parameter \( M \), we opt for an automatic approach to determine the best value for \( M \). Considering a point cloud composed of \( N \) points in three-dimensional space, with \( M \) as a variable from the set of natural numbers \( N \), each point \( p_{n} = (X, Y, Z)^T \) lies in \( \mathbb{R}^{3} \), and its \( M \)-neighborhood defines the extent of each receptive field \( \mathcal{R}_{n}^{M} \). To capture the local structure around each point \( p_{n} \), we compute the 3D covariance matrix, also referred to as the structure tensor \( S \), which is a symmetric positive definite matrix in \( \mathbb{R}^{3 \times 3} \). This tensor always has three non-negative eigenvalues \( \lambda_{1}, \lambda_{2}, \lambda_{3} \in \mathbb{R} \) that are ordered such that \( \lambda_{1} \geq \lambda_{2} \geq \lambda_{3} \geq 0 \) and correspond to an orthogonal set of eigenvectors. To determine the optimal neighborhood size for each point \( p_{n} \),  eigenvalues representing the principal components \cite{entropy} of the 3D covariance ellipsoid are obtained. The neighborhood sizes \( M \) of $p_{n}$ is optimized by minimizing the entropy of these eigenvalues, known as eigenentropy, defined as \cref{eq2}, which serves as a quantification of the spatial coherence or variability within the ellipsoid.

\begin{equation}
E_{\lambda}= - (e_{1}\ln(e_{1}) + e_{2}\ln(e_{2}) + e_{3}\ln(e_{3}))
\label{eq2}
\end{equation}

where $e_{j}=\lambda_{j}/\sum\limits_{i=1}^3 \lambda_{i}, (j = 1, 2, 3) $, representing each normalized eigenvalue. Specifically, we measure the eigenentropy in $[M_{\text{min}}, M_{\text{max}}]$, indicating the number of potential neighbors of $p_{n}$, and all integer values within this range are considered.\\

\subsection{Structural Convolution Operation}

Using the specified inputs and kernel definitions for 3D point cloud data, we propose the 3D Structural Convolution process, as shown in \cref{Fig3}. We determine the similarity between the receptive field $R^{M}_{n}$ and the direction-based kernel $K^{S}_{dir}$, represented as $Conv_{dir}(R^{M}_{n}, K^{S}_{dir})$, to derive point-specific features through cosine similarity. Additionally, we calculate the product of the distance to the farthest neighbor in $R^{M}_{n}$ and the weights in $K^{S}_{dist}$, represented as $Conv_{dist}(R^{M}_{n}, K^{S}_{dist})$, to obtain point-specific local features via Euclidean distance. These extracted features are then concatenated and processed through an MLP, enabling accurate feature extraction of the structural context without introducing bias towards either distance or directional context during training.\\

Unlike 2D CNNs, where kernels and image patches both have grid structures, our method involves combinations of 3D vectors instead of grids. For the purpose of convolution in 3D point clouds, we achieve our structural convolution operation with direction-based kernels and distance-based kernels. Initially, we assess the similarity between features within the receptive field of $p_{n}$ (i.e., $F(p_{n})$, $F(p_{m})$ for all $p_{m} \in N(p_{n}, M)$, as specified in \cref{eq1}) and the weight vectors of the directional kernel $K^{S}_{dir}$, centered around $k^{C}_{dir}$ with $S$ supports (namely, $w(k^{c}_{dir})$, $w(k^{s}_{dir})$ for all $s = 1, 2, \ldots, S$). We consider every pairing of $(p_{m}, k^{s}_{dir})$. Consequently, the direction-based convolution between a receptive field and a direction-based kernel can be described as follows:

\begin{equation}
\small
\begin{split}
Conv_{dir}(R^{M}_{n}, K^{S}_{dir}) &= \langle F(p_n), w(k^{c}_{dir}) \rangle + \sum\limits_{s=1}^{S} \max\limits_{m \in (1,M)} sim(p_{m}, k^{s}_{dir})
\end{split}
\label{eq3}
\end{equation}

where the symbol $\langle \cdot \rangle$ denotes the inner product operation, and the function $sim$ calculates the inner product between the features \(F(p_m)\) and the directional weights \(w(k^{s}_{dir})\), utilizing cosine similarity \cite{gcn} to define this interaction:

\begin{equation}
sim(p_{m}, k^{s}_{dir}) = \langle F(p_{m}), w(k^{s}_{dir}) \rangle 
\frac{\langle d_{m,n}, k^{s}_{dir} \rangle}{||d_{m,n}||\,||k^{s}_{dir}||}
\label{eq4}
\end{equation}

Moreover, to account for the influence of spatial relationships among neighboring points on structural characteristics, we assess the Euclidean distances from point $p_{n}$ to its most distant neighbor and then multiply them with the weights of the kernel $K^{S}_{dist}$ centered at $k^{C}_{dist}$ (precisely, $w (k^{c}_{dist}), w (k^{s}_{dist}), \forall s = 1, 2, \ldots, S$). we consider all conceivable combinations of  $(p_{m}, k^{s}_{dist})$ to perform the distance-based convolution that merges the a receptive field with the kernel, expressed as:

\begin{equation}
\small
Conv_{dist}(R^{M}_{n}, K^{S}_{dist})= \sum_{s=1}^{S}w^{s}_{dist}\times\max\limits_{m\in(1,M)}(||d_{m,n}||)
\label{eq5}
\end{equation}

Note that, since \cref{eq3} indicates the convolution between a receptive field and a direction-based kernel, and \cref{eq5} indicates the convolution between a receptive field and a distance-based kernel, the convolution operations in \cref{eq3} and \cref{eq5} must be applied to all receptive fields, all direction-based kernels, and all distance-based kernels.
Finally, by concatenating the outputs from the above convolution operations and passing them through an MLP, 3D structural convolution is achieved as follows:

\begin{align}
\label{eq6}
& Str-Conv(R^{M}_{n}, K^{S}_{dir}, K^{S}_{dist}) \\
&= MLP(Conv_{dir}(R^{M}_{n}, K^{S}_{dir})\oplus Conv_{dist}(R^{M}_{n}, K^{S}_{dist}))  \nonumber
\end{align}

Concatenating distance-based features and direction-based features, and then processing them through a Multi-Layer Perceptron (MLP), enables learning without bias towards either direction-based or distance-based kernels.

\begin{figure*}[!t]
    \centerline{\includegraphics[width=\textwidth]{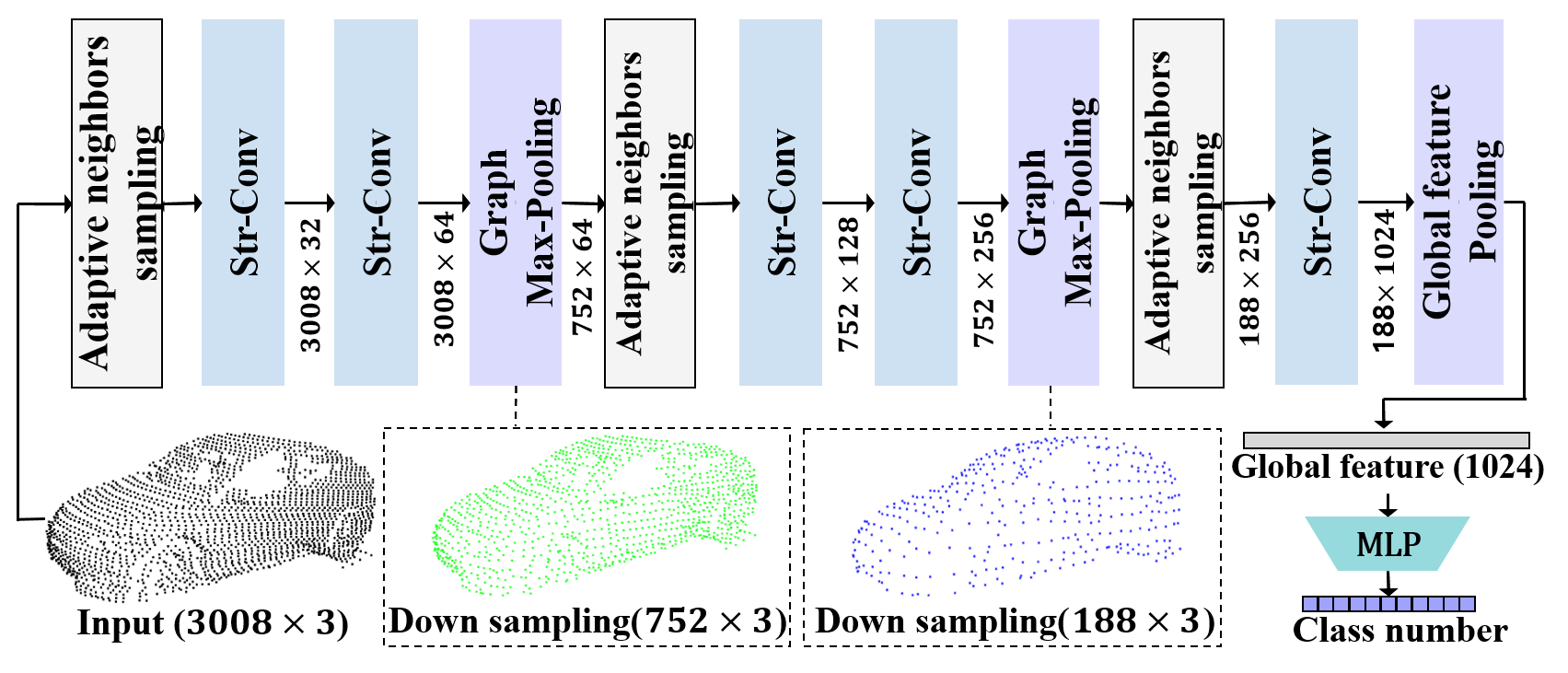}}
    \caption{Illustration of our classification architecture: Str-Conv blocks perform structural convolution. Graph Max-Pooling blocks perform channel-wise max-pooling from the features and then randomly sample a subset following the sampling rate $r$. During this process, the Adaptive neighborhood sampling block finds the optimal neighborhoods and passes them to the Str-Conv block.}
    \label{Fig4} 
\end{figure*}

\subsection{Classification Architecture}

In the design of the 3D-ASCN, we have modified the convolution and pooling layers. The framework consists of five convolution layers and three max-pooling layers, with adjustments made to the number of parameters to suit our needs for LiDAR-based point cloud data, as illustrated in \cref{Fig4}. In this model, the computation of the convolution layer follows the \cref{eq6} previously mentioned. First, the range of the neighborhood is used to be searched as input. The convolution process then uses a fixed number of kernels, each with a specific number of supports. These kernels are applied to a 3D point cloud and its associated D-dimensional features, generating output features for each network layer. The feature value for each point in the point cloud and the kernel weights are set to one. \\

Following the convolution step, the 3D Max-Pooling layer identifies the receptive field for each point and performs channel-wise max-pooling to compile features. It then samples a portion of the point cloud, effectively reducing the point count by a rate of 4. This pooling layer is crucial for efficiently learning features from the 3D point cloud by decreasing the dimensionality of both the input and output point clouds and their features by a quarter.


\begin{table}[t]
  \centering
  {\small{
    \begin{tabular}{c||c|c }
    \hline
    Train/Test Dataset   &   Method & Acc.(\%)\\
    \hline\hline
      & 3D-ASCN(Ours) &  99.2\\ 
      &    PointNet \cite{pointnet} &  99.5\\ 
       KITTI   &   DGCNN \cite{dgcnn}&  99.4\\ 
       (German, 64CH)    &   PointMLP \cite{pointMLP}&  \textbf{99.9}\\ 
           &   PointNeXt \cite{pointNext}&  99.6\\ 
    \hline
      & 3D-ASCN(Ours) & 92.2\\
      &   PointNet \cite{pointnet}&92.5\\ 
       nuScenes   &    DGCNN \cite{dgcnn}&  \textbf{95.9}\\ 
       (US, 32CH)       &   PointMLP \cite{pointMLP}&  94.7\\ 
           &   PointNeXt \cite{pointNext}&  95.5\\ 
          \hline
      & 3D-ASCN(Ours) & 94.2\\ 
      &   PointNet \cite{pointnet}& 92.9\\ 
       PanKyo   &    DGCNN \cite{dgcnn}& 96.3\\ 
       (Korea, 128CH)       &   PointMLP \cite{pointMLP}&  \textbf{96.6}\\ 
           &   PointNeXt \cite{pointNext}&  93.6\\ 
    
     \hline
    \end{tabular} 
  }}
  \caption{Intra-domain classification results on KITTI \cite{KITTI}, nuScenes \cite{nuScene}, and PanKyo \cite{PanKyo} datasets. The best results are given in \textbf{boldface}.}
  \label{table1}
\end{table}

\begin{table}
  \centering
  {\resizebox{\textwidth}{!}{
    \begin{tabular}{c|c||c|c||c|c||c|c }
    \hline
    Train  & Test  &  Method & Acc.(\%) & Train  & Test   &  Method & Acc.(\%)\\
    \hline\hline
    & & 3D-ASCN(Ours) & \textcolor{blue}{\textbf{79.3}} &  & &3D-ASCN(Ours) &  \textbf{89.8}\\ 
       &    & PointNet \cite{pointnet}&  54.8 &   &  &PointNet \cite{pointnet}& 6.1\\   
      KITTI &  nuScenes  & DGCNN \cite{dgcnn}&  49.7 &PanKyo & KITTI   & DGCNN \cite{dgcnn}&  14.7\\ 
        &   & PointMLP \cite{pointMLP}& \textbf{80.4} &   & & PointMLP \cite{pointMLP}& \textcolor{blue}{\textbf{87.7}} \\
        &   & PointNeXt \cite{pointNext}& 31.4  &   & & PointNeXt \cite{pointNext}& 5.8\\ 
    \hline
     & & 3D-ASCN(Ours) & \textbf{92.6} & &   & 3D-ASCN(Ours) &  \textbf{74.9}\\ 
          &   &  PointNet \cite{pointnet}& 40.4 & &   &  PointNet \cite{pointnet}& \textcolor{blue}{\textbf{44.4}}\\  
     nuScenes & KITTI  & DGCNN \cite{dgcnn}&  \textcolor{blue}{\textbf{66.3}}&  nuScenes & PanKyo  & DGCNN \cite{dgcnn}&  17.4 \\ 
          &   & PointMLP \cite{pointMLP}& 47.1 & &   & PointMLP \cite{pointMLP}&  15.2\\ 
          &   & PointNeXt \cite{pointNext}& 31.6 & &   & PointNeXt \cite{pointMLP}& 18.6\\ 
    \hline
     &   & 3D-ASCN(Ours) &  \textcolor{blue}{\textbf{55.6}} & && 3D-ASCN(Ours) &  \textbf{71.4}\\ 
     &   &  PointNet \cite{pointnet}& 17.6 &  &   &  PointNet \cite{pointnet}& 24\\  
     KITTI & PanKyo  & DGCNN \cite{dgcnn}&  19.9 & PanKyo  & nuScenes  & DGCNN \cite{dgcnn}&  46.2\\ 
          &   & PointMLP \cite{pointMLP}& 17.7 & &   & PointMLP \cite{pointMLP}& \textcolor{blue}{\textbf{64}}\\ 
          &   & PointNeXt \cite{pointNext}& \textbf{67.6} &  & & PointNeXt \cite{pointNext}& 17.3\\ 
     \hline
    
    \end{tabular} 
        }}
  \caption{Cross-domain classification results on KITTI \cite{KITTI}, nuScenes \cite{nuScene}, and PanKyo \cite{PanKyo} datasets. The bold \textbf{black}, \textcolor{blue}{\textbf{blue}} indicate the first and the second best performance on each scenario, respectively.}
  \label{table2}
\end{table}

\section{EVALUATION}
\label{section4}
The proposed algorithm is evaluated through comparisons with PointNet \cite{pointnet} DGCNN \cite{dgcnn}, pointMLP \cite{pointMLP}, and PointNeXt \cite{pointNext}. The newtwork is evalauted through the object classification accuracy and the evaluation metric is shown as follows:
$$
\text { Accuracy(\%) }=\frac{\text { the \# of correctly classified classes }}{\text { the \# of total ground truths }}\times100
$$

\subsection{Implementation Detail}

 Our model incorporates three pivotal components, including adaptive neighborhood sampling, direction-based kernel, and distance-based kernel. In the beginning,  the 10 nearest neighbors is sampled for each point using the kNN algorithm \cite{knearest}. The 3D-ASCN then evaluates the eigenentropy for each possible neighborhood size, from 3 to 10, to select the optimal number of neighbors. To accommodate kernels requiring a consistent size, the maximum neighborhood size previously determined is used. For cases with fewer than 10 neighbors, blanks are processed as the central location to ensure uniformity in input size, with the distance and cosine similarity of the central point set to zero. The 3D-ASCN applies both direction-based and distance-based kernels across all layers, with each layer featuring a fully connected (FC) layer for extracting distance features and another for direction features. This setup includes stacking five convolutional layers in a manner akin to 3D-GCN \cite{gcn}.
 
Experiments are conducted with our model on three outdoor point cloud datasets: KITTI \cite{KITTI}, nuScenes \cite{nuScene}, and PanKyo \cite{PanKyo}. These datasets are provided in the given link\footnote{https://sites.google.com/site/cvsmlee/dataset} and originate from various countries and manufacturers.  This diversity is crucial for assessing not only the model's inferencing capabilities but also its robustness across multiple domain shifts. Especially, they all have different channels. nuScenes has 32 channels, KITTI has 64 channels, while PanKyo has 128 channels. We evaluated the robustness of our model using these channel shifts. In addition, to seamlessly align the existing classes, we reduce the number of classes to three: Pedestrian, Car, and Truck.

\subsection{Classification on single-domain}

A detailed evaluation was conducted to assess the algorithm's performance, particularly its behavior under conditions that might lead to overfitting, by examining it from within-domain perspectives. \cref{table1} displays the classification performance within a singular domain. While 3D-ASCN shows a tendency towards underfitting in the general automotive dataset, a consequence of removing scale invariance and possessing low inductive bias, 3D-ASCN demonstrates performance on par with models known to overfit in specific domains, such as PointNet, DGCNN, pointMLP, and pointNeXt.

\begin{table}[t]
  \centering
  {\small{
    \begin{tabular}{c||c }
    \hline
    Method   &   Average Acc.(\%)\\
    \hline\hline
      3D-ASCN (ours) &  \textbf{77.3}\\
    \hline
      PointNet \cite{pointnet} &  31.2\\
    \hline
      DGCNN \cite{dgcnn}&  35.7\\ 
    \hline
      PointMLP \cite{pointMLP}&  \textcolor{blue}{\textbf{52}}\\
    \hline
      PointNeXt \cite{pointNext}&  28.7\\ 
    \hline
      
    \end{tabular} 
  }}
  \caption{Average accuracy for the cross domain scenarios of \cref{table2}. The bold \textbf{black}, \textcolor{blue}{\textbf{blue}} indicate the first and the second best performance on each scenario, respectively.}
  \label{table3}
\end{table}

\subsection{Channel shift robustness}

An evaluation was conducted across different domains to assess the models' robustness in various contexts. The experiments specifically aimed to examine the performance and robustness of models by training them in one domain and testing them in another, with a focus on domain invariance. These findings offer crucial insights into the models' performance across various datasets, facilitating a comprehensive assessment of their effectiveness.

\cref{table2} shows the performance comparison between 3D-ASCN and other methods for the cross-domain classification task of autonomous driving datasets. Our method demonstrates classification performance that consistently ranks among the top two across all domain shift scenarios. It especially outperforms the second-best methods by 26.3\% when shifting from nuScenes to KITTI, by 30.5\% when transitioning from nuScenes to Pankyo, and by 7.4\% when moving from Pankyo to nuScenes. Moreover, the average accuracy of 3D-ASCN surpasses the second-best method, pointMLP, by 25.3\%, as illustrated in \cref{table3}.

Consequently, Both the kernels based on cosine similarity, the kernels based on Euclidean distance, and adaptive neighborhood selection methods were effective in enabling structural learning capabilities amid general changes in datasets.

\subsection{Ablation study}

The key to Structural Convolution is to extract features of 3D structural context by combining Cosine similarity and Euclidean distance terms and, through this, ensure that the model has robustness in situations where the LiDAR sensor changes. For this purpose, we considered three methods of convolution operations, as shown in \cref{table4}. When using only direction-based kernels, performance deteriorates in situations where the Lidar sensor is changed. In contrast, the robustness of the model can be obtained by using both direction-based kernels and distance-based kernels. Moreover, Concatenating distance-based features and direction-based features, and then passing them through a Multi-Layer Perceptron (MLP), enables learning without bias towards either direction-based or distance-based kernels. Therefore, our structural convolution achieved the best performance both in situations where the domain was constant and in situations where the domain changed.

\cref{table5} illustrates the impact of adaptive neighborhood sampling. Specifically, we set the fixed $M$ to three. The network's performance improves by selecting the number of neighbors adaptively, particularly in situations involving domain changes, showing performance enhancement. When trained and tested within the same domain, such as both being in Pankyo, the performance was the same because the optimal number of neighbors was close to three. These results demonstrate that adaptive $M$ generally offers great utility in scenarios involving cross-domain changes.

\begin{table}
  \centering
  {\small{
    \begin{tabular}{c|c|c|c}
    \hline
    Train & Test   &   Convolution operation & Acc.(\%)\\
    \hline\hline
    & \multirow{3}{*}{KITTI} & $Conv_{dir}$ & 52.6 \\
    & & $Conv_{dir}+Conv_{dist}$& 86.3 \\
    & & $Str-Conv$ & \textbf{88.7} \\ \cline{2-4}
    \multirow{3}{*}{PanKyo}& \multirow{3}{*}{nuScenes} & $Conv_{dir}$ & 55.7 \\
    & & $Conv_{dir}+Conv_{dist}$ & 69.0 \\
    & & $Str-Conv$ & \textbf{71.2} \\ \cline{2-4}
    & \multirow{3}{*}{PanKyo} & $Conv_{dir}$ & 92.4 \\
    & & $Conv_{dir}+Conv_{dist}$ & 93.9 \\
    & & $Str-Conv$& \textbf{94.2} \\ \hline
    \end{tabular}
  }}
  \caption{Performance with different convolution operations on  KITTI \cite{KITTI}, nuScenes \cite{nuScene}, and PanKyo \cite{PanKyo} datasets. The best results are given in \textbf{boldface}.}
  \label{table4}
\end{table}

\begin{table}
  \centering
  {\small{
    \begin{tabular}{c|c|c|c}
    \hline
    Train & Test  & The number of neighbors & Acc.(\%)\\
    \hline\hline
    & \multirow{2}{*}{KITTI} & Fixed M & 88.7 \\
    & & Adaptive M & \textbf{89.8}\\ \cline{2-4} 
    \multirow{3}{*}{PanKyo} & \multirow{2}{*}{nuScenes} & Fixed M & 71.2 \\ 
    & & Adaptive M & \textbf{71.4}\\ \cline{2-4}
                            & \multirow{2}{*}{PanKyo} & Fixed M & \textbf{94.2} \\ 
    & & Adaptive M & \textbf{94.2}\\ \hline
    \end{tabular}
  }}
  \caption{Performance comparison according to neighborhood sampling methods on  KITTI \cite{KITTI}, nuScenes \cite{nuScene}, and PanKyo \cite{PanKyo} datasets. The best results are given in \textbf{boldface}.}
  \label{table5}
\end{table}

\section{CONCLUSION}
\label{section5}
In this study, we introduce the 3D-ASCN, a novel model designed for feature extraction with minimal inductive bias. The 3D-ASCN framework incorporates a Distance Feature Extraction technique alongside an Adaptive Nearest Neighbor approach, strategically devised to mitigate the scale-invariant bias inherent in the GCN. Consequently, while 3D-ASCN demonstrates modest inference performance within a singular domain, it exhibits enhanced robustness against domain shifts, including channel reduction. This adaptability is particularly significant in addressing the sensor-dependency challenge prevalent in point cloud classification tasks, thereby potentially advancing the versatility of autonomous driving technologies. In light of these findings, our future research endeavors will focus on exploring efficient learning methodologies utilizing small datasets. Specifically, we aim to investigate self-supervised learning paradigms based on raw point clouds, with the objective of achieving superior performance within individual domains.



\begin{thebibliography}{999}



\bibitem{review1}
Kuutti, Sampo, et al. "A survey of deep learning applications to autonomous vehicle control." IEEE Transactions on Intelligent Transportation Systems 22.2 (2020): 712-733.
\bibitem{review2}
Li, Ying, et al. "Deep learning for lidar point clouds in autonomous driving: A review." IEEE Transactions on Neural Networks and Learning Systems 32.8 (2020): 3412-3432.

\bibitem{pointnet}
Charles Ruizhongtai Qi, Hao Su, Kaichun Mo, and Leonidas J. Guibas. Pointnet: Deep learning on point sets for 3d classification and segmentation. In Proceedings of the IEEE conference on computer vision and pattern recognition (CVPR), pages 652–660, 2017.

\bibitem{pointnet+}
Charles Ruizhongtai Qi, Li Yi, Hao Su, and Leonidas J. Guibas. Pointnet++: Deep hierarchical feature learning on point sets in a metric space. In Advances in neural information processing systems (NIPS), volume 30, 2017

\bibitem{dgcnn} Yue Wang, Yongbin Sun, Ziwei Liu, Michael M Bronstein Sanjay E Sarma, and Justin M Solomon. Dynamic graph cnn for learning on point clouds. Acm Transactions On Graphics (tog), 38(5):1–12, 2019.

\bibitem{p8}  Qiangui Huang, Weiyue Wang, and Ulrich Neumann. Recurrent slice networks for 3d segmentation of point clouds. In Proceedings of the IEEE Conference on Computer Vision and Pattern Recognition (CVPR), 2018.


\bibitem{p5}  Matheus Gadelha, Rui Wang, and Subhransu Maji. Multiresolution tree networks for 3d point cloud processing. In Proceedings of the European Conference on Computer Vision (ECCV), 2018.


\bibitem{p10} Roman Klokov and Victor Lempitsky. Escape from cells: Deep kd-networks for the recognition of 3d point cloud models. In Proceedings of the IEEE international conference on computer vision (ICCV), 2017.


\bibitem{gcn} Zhi-Hao Lin, Sheng-Yu Huang, and Yu-Chiang Frank Wang. Learning of 3d graph convolution networks for point cloud analysis. IEEE Transactions on Pattern Analysis and Machine Intelligence, 44(8):4212–4224, 2021.




\bibitem {Scaleselection} Jean-François Lalonde, Ranjith Unnikrishnan, Nicolas Vandapel, and Martial Hebert. Scale selection for classification of point-sampled 3d surfaces. In Proceedings of the International Conference on 3-D Digital Imaging and Modeling (3DIM), pages 285–292, 2005.


\bibitem {curvature} Mark Pauly, Richard Keiser, and Markus Gross. Multi-scale feature extraction on point sampled surfaces. Computer Graphics Forum, 22(3):81–89, 2003.


\bibitem {dimensionality} Jérôme Demantké, Clément Mallet, Nicolas David, and Bruno Vallet. Dimensionality based scale selection in 3d lidar point clouds. Int. Arch. Photogramm. Remote Sens. Spatial Inf. Sci., XXXVIII(5/W12):97–102, 2011.

\bibitem {entropy} Martin Weinmann, Boris Jutzi, and Clément Mallet. Semantic 3d scene interpretation: A framework combining optimal neighborhood size selection with relevant features. ISPRS Ann. Photogramm. Remote Sens. Spatial Inf. Sci., II(3):181–188, 2014.


\bibitem {cylinder} Sagi Filin and Norbert Pfeifer. Neighborhood systems for airborne laser data. Photogrammetric Engineering Remote Sensing, 71(6):743–755, 2005.



\bibitem {sphere} Impyeong Lee and Anton F Schenk. Perceptual organization of 3d surface points. International Archives of the Photogrammetry, Remote Sensing and Spatial Information Sciences, XXXIV(3A):193–198, 2002.

\bibitem {knearest}Lars Linsen and Hartmut Prautzsch. Local versus global triangulations. In Proceedings of Eurographics, pages 257–263, 2001.


\bibitem{KITTI}
Andreas Geiger, Philip Lenz, and Raquel Urtasun. Are we ready for autonomous driving? the kitti vision benchmark suite. In 2012 IEEE Conference on Computer Vision and Pattern Recognition. IEEE, 2012.

\bibitem{nuScene}
Holger Caesar, Varun Bankiti, Alex H Lang, Sourabh Vora, Venice Erin Liong, Qiang Xu, Anush Krishnan, Yu Pan, Giancarlo Baldan, and Oscar Beijbom. nuscenes: A multimodal dataset for autonomous driving. In Proceedings of the IEEE/CVF conference on computer vision and pattern recognition, 2020.

\bibitem{pointMLP}
Xu Ma, Can Qin, Haoxuan You, Haoxi Ran, and Yun Fu. Rethinking network design and local geometry in point cloud: A simple residual mlp framework. In International Conference on Learning Representations (ICLR), 2022.

\bibitem{ModelNet40}
MZhirong Wu, Shuran Song, Aditya Khosla, Fisher Yu, Linguang Zhang, Xiaoou Tang, and Jianxiong Xiao. Shapenets: A deep representation for volumetric shapes. In Proceedings of the IEEE Conference on Computer Vision and Pattern Recognition (CVPR), 2015.

\bibitem{ScanObjectNN}
Mikaela Angelina Uy, Quang-Hieu Pham, Binh-Son Hua, Duc Thanh Nguyen, and Sai Kit Yeung. Revisiting point cloud classification: A new benchmark dataset and classification model on real-world data. In International Conference on Computer Vision (ICCV), 2019.

\bibitem{pointNext}
Guocheng Qian, Yuchen Li, Houwen Peng, Jinjie Mai, Hasan Abed Al Kader Hammoud, Mohamed Elhoseiny, and Bernard Ghanem. Pointnext: Revisiting pointnet++ with improved training and scaling strategies. arXiv preprint arXiv:2206.04670, 2022.


\bibitem{PanKyo}
Rohee Lee, Seonghoon Ryoo, and Soomok Lee. Domain-invariant 3d structural convolutional network for autonomous driving point cloud dataset. In IEEE Intelligent Vehicles Symposium (IV), 2024.

\end{thebibliography}
\end{document}